\documentclass[11pt]{article}

\usepackage{mdframed}
\usepackage{enumitem}
\usepackage{dblfloatfix}
\usepackage[nohyperref]{acl2018}
\usepackage{tasklib}
\usepackage{zensec}
\usepackage{zenmath}
\usepackage{zenacademic}
\usepackage{zenhref}
\usepackage{zentable}
\usepackage{zenfig}
\usepackage{zennlp}

\usepackage{tasktables}

\zstwotitlesameinstacl{Barbara Di Eugenio}{bdieugen@uic.edu}
\aclfinalcopy

\begin{document}
\maketitle
\begin{abstract}
\Todolists\ are a popular medium for personal information management. As
\todos\ are increasingly tracked in electronic form with mobile and desktop
organizers, so does the potential for software support for the corresponding
tasks by means of \agentslong.  While there has been work in the area of
\pas\ for \todos, no work has focused on classifying user intention and
information extraction as we do.  We show that our methods perform well across
two corpora that span sub-domains, one of which we released.
\end{abstract}


\zzsec{todo-intro}{Introduction}

\Todolists\ are pervasive and offer a concise representation of tasks that
need to be accomplished~\cite{bellotti2004,gil2012capturing}, with studies
showing that up to 60\% of the population use them~\cite{jones1997}.
\Todos\ are often expressed as short utterances and written in list form.
Examples of \todo\ are \tdbuyex, \tdcallex, and \td{Hotel reservation}.

With the growing popularity of \pas, such as Apple's voice-based Siri, the need
for understanding and executing tasks such as those embodied in
\todos\ continues to grow; conversely, there is great potential in automating
the resolution of \todos\ given the plethora of mobile and desktop applications
that currently exist. 

This paper focuses on describing several models we have constructed to
interpret \todos. As far as we know, there has been little research
in parsing and classifying \todos, which entirely consists of short requests
couched either as imperative sentences or as fragments not containing a verb;
neither imperatives nor fragments are as frequent in corpora as other forms in
which a user's intention can be expressed, as we will discuss in
Section~\ref{related-work}.

Execution of these tasks can be achieved by invoking an appropriate \agent,
which will act upon the task~\cite{gil2012capturing}.  For example, \tdcallex,
would yield: \tdcallexparsed\ and would look up {\tt mom} as contact and dial
via the phone application.  \Agents\ are described in \zzsecref{corpus}.

The challenges with processing \todos\ are:
\begin{itemize}
\setlength\itemsep{.3em}
\setlength{\parskip}{0pt}
\setlength{\parsep}{0pt}
\item {\bf Short length of utterances}, which yield poor results with current
  methods~\cite{han2011lexical}.

\item{\bf Missing head verb}, which makes classification more difficult if
  missing.  For example, \td{Hotel reservation} lacks the head verb,
  {\it schedule}.

\item{\bf Disambiguation of named entities}, which include products to
  purchase, persons with whom to communicate, companies, etc.

\item{\bf Processing commands in imperative form}, for which in most systems
  are well formed and don't contain many named entities.
\end{itemize}

In this paper, we discuss our approach to mapping a \todo\ to a corresponding
\agent; this includes extracting its \arguments\ and classifying them as
concerns their types. We will describe the corpus we built, the pre-processing
steps we devised, and the features we used to inform our classifiers.

Finally, we will demonstrate the generality of our approach by applying it to a
compiled and annotated corpus with a partially separate domain.


\zzsec{related-work}{Related Work}

\label{related-work}

From a linguistic point of view, \todos\ are either expressed with (short)
imperative sentences or with fragments that do not contain a verb. Imperatives
have not been studied very extensively in NLP, since they (not surprisingly) do
not occur frequently in corpora of ``standard'' English, such as newspaper
articles.  For example, the PARC 700 dependency bank~\ci{king_parc_2003} is a
random sample of 700 utterances taken from the Wall Street Journal
corpus~\ci{marcus1994penn} and contains only seven utterances containing at
least one imperative.

However, even in datasets that focus on dialogue and tasks to be performed,
imperatives do not occur that often. To start with,
as~\ci{levinson1983pragmatics} notes, the imperative is rarely used to issue
requests in English. This seems to be borne out even in interactions with \pas,
even if, beyond anectotal evidence, distributional analyses of such data are
not widely available.

One such analysis is provided by \ci{tur2014detecting}, who analyzes
interactions between users and an entertainment \pas. They found that 31.21\%
of such utterances start with a VB tag (an upper bound for imperatives),
another 31.64\% with tags representative of NPs and wh-NPs (the remaining 37\%
of utterances is left unspecified): so even when communicating with an
entertainment \pas, users use imperative sentences only about $\frac{1}{3}$ of
the time (this frequency is not rare, but much lower than one would expect
given the circumstances).  In their work on evaluating the performance of \pas\
such as Siri, Google Now and Cortana, Jiang et
al.~\shortcite{jiang_automatic_2015} show that 67\% of user action types are
commands, but they don't specify the proportion of imperatives within those.

A substantial percentage of \todos\ are expressed as fragments, such as
\td{Hotel reservation}. The literature on fragments is also rather sparse,
other than as concerns ellipsis \ci{kempson2015}--but it is not clear that
these fragments are in fact ellipsis. Even tweets, the shortest of today's
social media language, are often grammatically ill-formed, contain a main verb.

If we now turn to the interpretation of \todos, Gil et
al.~\shortcite{gil2012capturing} explored different kinds of intelligent
assistance for \todolists.  Gil et al.~\shortcite{gil2012capturing} provides a
manual data entry to categorize the \todo\ and Jiang et
al.~\shortcite{jiang_automatic_2015} yields a performance evaluation, however,
neither automatically categorize the \agent\ as we do. Likewise
\ci{tur2014detecting} distinguishes between requests that are covered by the
current interpreter and those that are not, but does not interprets them in
anyway.

\zztagentdist

In this work, we also tackle \argument\ extraction. \footnote{Argument
  extraction can be considered equivalent to slot filling as defined in many
  spoken dialogue systems.  We follow e.g. Propbank~\ci{kingsbury2002treebank}
  in using the term ``\arguments''.} Our method of extracting verbs and their
\arguments\ is similar in spirit to efforts like
OpenIE~\cite{fader2011identifying} and NeLL~\cite{NELL-aaai15}.  Our work is
specifically targeted at \todolists\ where sentences are very short and
telegraphic.  OpenIE and NeLL learn from a large repository of knowledge, whose
language is not telegraphic, and mostly well formed.

Earlier work in \argument\ extraction includes processing sequences of
executable actions for the Windows operating system by Branavan et
al.~\shortcite{branavan2009reinforcement}.  Other \argument\ extraction work
includes efforts in deep learning by Meerkamp et
al.~\shortcite{meerkamp2017boosting} for boosting precision of existing
extraction systems.  However, our work covers a broader domain of general
English \todos\ and does not fit in any existing method of tagging \arguments.

The approach taken by Ghani et al.~\shortcite{ghani2006text} is the most
similar to ours with regards to their \argument\ extraction for textual product
descriptions.  However, their \argument\ set is restricted to a predefined set
where our model learns the \arguments\ to extract.

\zztargdist


\zzsec{corpus}{Corpora}

This work involves two \todo\ data sets: the first is a proprietary
corpus\footnote{Regretfully, we can not share the proprietary annotated
  corpus.} and the second is a publicly available corpus we have built and
released\footnote{\url{\corpusurl}}.

The corpus consists of \agent\ and \argument\ annotations.  The \agent\ type
was formulated from an initial analysis of the corpus and several iterations of
annotation guideliness that led us to the final list given
in~\zztabref{agent-dist}.  The \argument\ annotations are tokens associated
with an \agent\ annotation and provide additional context helpful in task
resolution.  A complete list of \arguments\ are given in~\zztabref{arg-dist}
and described in more detail in~\zzsecref{arg-anon}.

\zzsubsec{proprietarycorp-intro}{\Thedataset}

\Thedataset\ was collected from two sources: publicly available online sources
(see Appendix~\ref{sec:corp-sources} for a full list of sources) and private
sector data.  We annotated a random sample of 3,169 \todos\ with one utterance
per task and doubly annotated 1,342 to compute the kappa score.  We then
divided the 3,169, into usable {\it non-exceptions} (1,690) and unusable {\it
  exceptions} (1,479) \todos.  A task was considered as an exception for one of
the following reasons:
\begin{itemize}
\setlength\itemsep{0em}
\renewcommand\labelitemi{\small$\bullet$}

\item the \todo\ itself is ambiguous (\ie \td{flowers}--plant them or buy them?)

\item language is not English (\ie \td{compra flores}) or meaningless (\ie
  \td{mkmkmk})

\item illegal activity (\ie\ \td{buy drugs})

\item generic professional (\ie \td{first quarter presentation})
\end{itemize}

The 1,690 non-exception tasks were annotated with \agent\ labels and their
respective \arguments.  \agentscap\ with less than 15 utterances were pruned as
anything less proved to be insufficient training data.  This left 1,611 tasks
with utterances used for training and testing.


\zzsubsec{pubcorp-intro}{\ThePubDataset}

We created a \pubdataset\ composed of 102 volunteer contributed personal
\todos\ and 498 Trello\footnote{\url{http://trello.com}} \todos\ with \agent\
annotations.  A subset of this data, including 68 volunteer and 218 Trello
scraped \todos, was used to test and train the model.  The Trello data was
sourced from public boards that allow for redistribution and the volunteers
agreed to release their data for public distribution.  This corpus does not
contain \argument\ annotations given the purpose was to provide a way to
reproduce \agent\ classification results.


\zzsubsec{arg-anon}{Argument Annotations}

There is a zero to many relationship between \agents\ and \arguments.  For
example, \td{grocery store} is tagged as a \texttt{buy} \agent\ with no
\arguments.

\zzfutterance

\zzTabref{arg-dist} lists the \argument\ annotations and their distribution in
the corpus.  Note that some \arguments\ span multiple \agents\ while others are
specific to a particular \agent.  An additional {\tt descriptor list}
annotation was provided to address edge cases where existing modifiers were
insufficient.  For example, the descriptor list would be populated with a URL
since there is no corresponding \argument\ for \texttt{calendar appointment}.

A fully annotated task of \tdbuysweater\ is shown in \zzfigref{utterance} and
has the annotation \tdbuysweaterparsed.

An inter-coder agreement metric was computed to get an idea of how consistent
the corpus was being annotated using Fleiss' kappa~\cite{fleiss1971measuring}.

Early efforts were made to create a consistent corpus with sufficient overlap
and annotation guidelines, in the face of subtle differences.  This work proved
worthwhile as the coder agreement of \agentclass\ produced an agreement 0.679
over 848 tasks annotated by two annotators.

The \pubdataset\ contains no argument annotations.


\zzsec{pipeline-processing}{Corpus Processing}

In earlier experiments, significant propagation errors in the pipeline were
found that resulted in poor \agent\ and \argument\ classification due to
incorrect \postagger\ \electedverbs\ (described in \zzsecref{agent-class}).  To
overcome, this we added a pre-processing step to enhance the \stanfordnlp\ and
created features from the parsed utterances to train and test the models.
This process includes:
\begin{enumerate}
  \item Extend the \stanfordner. First, we compiled named entity lists that are
    supplied to the \ner\ system at parse time and then we parsed the
    annotations.

  \item Build the \firstverbmodel. To ameliorate errors caused from issues of
    parsing short utterances, the \postagger\ was enhanced by creating the
    \firstverbmodel.
 
\item Apply the \firstverbmodel\ to the corpus. This prevents significant error
  propagation by correcting the first POS tag created in the previous step.
  This correction is seen later as \agent\ and \argument\ models utilize the
  \electedverb\ parsed at this step.
\end{enumerate}


\zzsubsec{ext-ner}{Extending Named Entity Recognition}
\label{sec:ner}

\Ner\ proved to be crucial as it provides additional context for
classification.  Two sets of features were created using both the
\stanfordnerintro\ and the \stanfordregintro.  The latter was enhanced to
include a set of static word lists (\nerlists) generated from \wikidataintro,
\opdintro, and \datalists.

The \wikidata\ lists were created with a set of SPARQL~\cite{prud2006sparql}
queries that included a primary term, and for some lists, a sublist term.
\Thedatalists\ also were annotated with three levels of categories, gender for
human names, and list type (\ie\ {\tt modifiers}, {\tt attributes} and {\tt
  products}).  These \nerlists\ were used to create a \stanfordreg\ formatted
regular expression input file and subsequently used during parse time for each
\todo.  The combined \wikidata\ and \datalist\ produced a total of 85,777
unique entities.

\zzsubsec{fvm-build}{Building the \Firstverbmodel}

The most crucial error made by the \postagger\ was incorrectly tagging the
first token of utterances as non-verbs.  The input to the \firstverbmodel\ are
features of the parsed utterances given by the \stanfordnlp\ and the class is
the \electedverb, which is used by the \agent\ and \argument\ models.  Out of
1,690 utterances, 653 (38.6\%) have the \electedverb\ annotation in the
\dataset, which is similar to the initial word utterance relative POS tag
frequences (31.21\%) of the Tur et al.~\shortcite{tur2014detecting} VPA and web
search datasets.

\zzsubsubsec{fvm-bootstrap}{Bootstrap the Model}

For the first parsing of the corpus this error was corrected by reassigning the
POS tag of the initial token using the following criteria:
\begin{enumerate}[label=\alph*\upshape)]
\item identified as a present tense verb tag in \wordnetintro\ and
\item identified as not a color as in ``{\it yellow} curry'' is a noun and not
  a verb as in ``{\it yellow} your teeth from coffee''.
\end{enumerate}


Each annotation included a token with a verb POS tag that differentiated each
\todo\ across \agents, which was in turn used as a feature.  This was used to
test the accuracy of the procedure to replace the POS tag with a verb tag.
However, this method proved to be detrimental for \todos\ containing homonyms.

\zzsubsubsec{fvm-build}{Build the Rule Based Model}

The Stanford POS tagger performed with an F-measure of 0.88.  A rule set
classifier~\cite{Frank1998} was created using \wekaintro\ brought the
F-measure up to 0.92 using the following features in addition to the
features created in bootstrap model described in \zzsecref{fvm-bootstrap}:
\begin{enumerate}[label=\alph*\upshape)]
\item the POS tags of the first token from both models:
  \begin{enumerate}
  \item[a$_1$)] the Stanford POS tag
  \item[a$_2$)] the \firstverbmodel\ POS tag~\zzseesec{fvm-bootstrap}
  \end{enumerate}
\item sentences containing one word
\item NER token spans greater than 1.
\end{enumerate}

After creating the rule based model as just described, we applied the rule
based model to correct the first POS tag of all of the utterances in the corpus
using the \stanfordnlp.  The input to this step are the utterances and
annotations from the corpus and the output is a tree of parsed items, which are
used as features for the \agent\ and \argument.

\zztsrlutterance


\zzsec{todo-model}{Build the Classification Models}

In this section we cover the techniques to process \todos, classify the \agent,
and perform \argex.

\zzsubsec{arg-extraction}{Argument Extraction}

The \param\ model is trained first as the \agent\ model uses its labels as
features.

A \srlerintro\ was used for many of the features in the \argument\ model.
There are one or more uses of a verb depending on the context, and each use of
the verb has its own argument set.  For this work, \propbankintro\ was used for
the operative verb of the utterance.

The SRL and POS tagging of the example given in \zzfigref{utterance} is shown
in \zztabref{srl-utterance}.  The {\tt head-dep} label and {\tt
function-tag} represent how a parent and child token node are related in
a head dependency tree~\cite{de2006generating}.  This example's columns are
further explained in \zztabref{param-feats}.

The \agent\ model was trained and tested (see \zzsecref{todo-results}) using
\weka\ on a per-\agent\ basis.  The \clearnlpintro\ semantic role labeling
parser that was trained on \propbank\ verb classes was used to provide features
to the \param\ model.  These feature's descriptions are listed in
\zzTabref{param-feats}.

All feature set permutations were used in a ten-fold cross validation with the
highest performing sets listed in \zzTabref{param-feat-sets}.  Results are
given by \agent\ in \zzTabref{param-results}, which do not include techniques
that failed (\ie\ raw word vectors for the word $w_n$).

\zztparamfeats

\zzsubsec{agent-class}{\agentsec\ Classification}

The \electedverb\ feature is the head root node of the dependency tree.  If
there is no parsed head node the POS tag of the first token of the utterance is
used.  For example, \tdbuyex\ correlates closely to the {\tt buy} \agent.  The
\electedverb\ was the first feature when creating the model, and by itself,
provided an impressive accuracy 55\% by itself.

\zztparamfeatsets

\zzagentresults

\zzpubagentresults

\zztparamresults

We used the lemmatized form of the token for word count and cosine similarity
features.  Let $c_{wa} = \mathtt{Count(w, a)}$ be the count of word $w$ for
\agent\ $a$ and $C_a$ be the set of word counts per \agent\ such that
$c_{wa} \in C_a$.  We limit $C$ to contain the highest $n$ frequency counts
with $n = |C_a|$ and hold $n$ constant for all \agents\ as a hyper parameter.
We use the word count aggregation across $C_a$ as feature:
\begin{equation}
WC_a = \sum_{c \in C_a} c
\label{eq:wc-feature}
\end{equation}

Significant performance gains were achieved by increasing $n$ from 5 to 15
with the $WC_a$ feature.  Now we define a mapping from word to a word
distribution over $C$ marginalizing over the word frequency:

\begin{equation}
q(w, a) = \frac{c_{wa}}{WC_a}
\label{eq:wc-dist}
\end{equation}

For example, for the \texttt{buy} \agent\ utterances \td{Purchase a shirt.
  Iron shirt.}:
$C_\textrm{buy} = \{ c_\mathrm{purchase}=1, c_\mathrm{a}=1 , c_\mathrm{iron}=1,
c_\mathrm{shirt}=2 \}$ and %
$q(\mathit{purchase}, \mathtt{buy}) = 1/4$,
$q(\mathit{a}, \mathtt{buy}) = 1/4$, $q(\mathit{iron}, \mathtt{buy}) = 1/4$,
$q(\mathit{shirt}, \mathtt{buy}) = 2/4$.

In addition, word vector cosine distance~\cite{mikolov2013efficient} was
calculated with \jwordvecintro\ using the the English Wikipedia pre-trained
word vector data set~\ci{mahoney2006rationale}.  The word vector library was
used by summing over the token cosine similarity and weighting it with the word
frequency distribution from equation~\ref{eq:wc-dist}.

The cosine similarity feature is created by calculating the MLE across all
\agents\ $A$ to create cosine similarity (CS) for each sentence $S$:

\begin{equation*}
\label{eq:cos}
CS_s = \zzargmax_{a \in A}\sum\limits_{w_c \in C_a}\sum\limits_{w \in S}q(w_c, a)
\cdot \mathtt{cos}(w_c, w_s)
\end{equation*}

This feature contributed to a 5 point increase in F-measure in all results
reported in \zzTabref{agent-results}.


\zzsec{example}{An Illustrative Example}

To illustrate how \todos\ are handled we will use the example \td{new christmas
  sweater for john}.  Once we receive the utterance the following happens:
\begin{enumerate}
\item Tokenize and sentence chunk, and POS tag the utterance using the modified
  version of the Stanford POS Tagger.
\item Create a head tree and tag tokens with \propbank\ data using the \srler.
\item Classify \agent\ as \td{buy} using the word counts, named entities and
  word vectors~\zzseesec{agent-class}.
\item For each token in the utterance using the parameter model for
  \agent\ \td{buy} classify an \argument\ ``sweater'' as \td{item}, ``john'' as
  \td{person} and ``christmas'' as \td{holiday}.
\item Concatenate contiguous tagged tokens of same \argument\ type.  In this
  example there are none, but if the example used ``blue sweater'', both would
  be tagged as \td{item} and be returned as one argument.
\end{enumerate}


\zzsec{todo-results}{Results}

Since testing was exhaustive, only noteworthy performance results for the
\agent\ and \argument\ models are given.  All subsets of reported features
along with hyperparameter tuning was tried.  The \agentclass\ and baseline
results are given in \zzTabref{agent-results} and \paramclass\ results are
given in \zzTabref{param-results}.  Note that the \paramclass\ results are for
\arguments\ and classify over all \agents, including the low count \agents\
shown in \zztabref{agent-dist}.

\zzsubsec{agent-res}{\agentsec\ Classification}

The baseline was created from the majority \agent\ class
\zzseetab{agent-results}.

A ten-fold cross validation was used on the \agent\ model.  Many classifiers
and feature combination sets were tested.  \logitboost\ had the highest
performance.  The $\tilde\chi^2$ was computed between all classifiers with \#7
showing a significant performance improvement increase over \#2 - \#4 with $p <
0.01$ in \zztabref{agent-results}.

Results from the \pubdataset~\zzseesec{pubcorp-intro} show a very similar
pattern to those of \dataset\ as shown in~\zztabref{pub-agent-results}.

\zzsubsec{param-res}{Argument Classification}

The \paramclass\ results are given in \zztabref{param-results} for each
respective \agent\ and show a wide F-measure variance.  The model was trained
and tested over a high variance of \argument\ occurrences as shown
in~\zztabref{agent-dist} with some \agents\ covering many more annotations than
others (\ie\ ``Buy'' was the majority \agent\ consisting of 28.5\% of the task
annotations).  Another reason for the wide distribution in results is the
ambiguous nature of some \agents.  For example, {\tt self-improve} could be
anything from {\it study}, {\it school work} or {\it physical exercise}.

\zzsec{conclusion}{Conclusion}

Classifying \todos\ with good performance from upstream parsed data is a
tractable problem.  Using Argument extraction to aide in automating \todo\
items is possible using the methods outlined in this work.


Bootstrapping methods for a NER with product lists using semi-supervised
methods was used by Putthividhya, Pew and
Junling~\shortcite{putthividhya2011bootstrapped}.  Similarly, there is
sufficient motivation by using our \nerlists\ for exploring generation of
entities using similar methods.

We focused on the \agent\ and \argument, but more work is needed to classify a
category of the task, which identifies the theme of the action or its product
attribute~\cite{ghani2006text} as a node in product taxonomy (\ie\ \td{buy
  dress} $\{dress\} \rightarrow (apparel, women)$).

\medskip
\bibliography{task}

\appendix

\zzsec{corpus-sources}{Corpus Sources}
\label{sec:corp-sources}

\Thedataset \todolist\ corpus was taken from the following locations:
\begin{itemize}
\small
\item \url{http://msippey.tadalist.com/lists/public/155420}
\item \url{https://wiki.itap.purdue.edu/display/INSITE/Ta-Da+List+Research}
\item \url{https://www.rememberthemilk.com/help/?ctx=basics.publish.publishlistpublic}
\end{itemize}

\noindent
The public \todolist\ corpus was taken from the following location:
\begin{itemize}
\small
\item \url{https://trello.com}
\end{itemize}

\end{document}